%% file: main.tex
\pdfoutput=1

\documentclass[11pt]{article}

\usepackage[preprint]{acl}

\usepackage{times}
\usepackage{latexsym}
\usepackage[T1]{fontenc}
\usepackage[utf8]{inputenc}
\usepackage{microtype}
\usepackage{inconsolata}
\usepackage{graphicx}

\usepackage{booktabs}
\usepackage{amsmath}
\usepackage{amssymb}

\title{Knowing When Not to Answer: Cross-Domain and Multi-Turn Generalization of Latent Underspecification Signals}

\author{
  Jerzy Kami\'nski \quad Ilya Galyukshev \quad Artem Kuznetsov \\
  \textbf{Danil Fedorov \quad Kirill Redko \quad Sergey Chuprin} \\
  \textbf{Aidar Shumbalov \quad Stanislav Chumakov \quad Anna Kalyuzhnaya} \\
  ITMO University, Saint Petersburg, Russia
}

\begin{document}

\maketitle

\input{sections/abstract}

\input{sections/introduction}

\input{sections/method}

\input{sections/results}

\input{sections/related_work}

\input{sections/conclusion}

\input{sections/limitations}

\bibliography{references}

\input{sections/appendix}

\end{document}

%% file: sections/abstract.tex
\begin{abstract}
Large language models routinely answer questions that cannot be answered from
the information given, and in dialogue they answer before enough has been said.
Unanswerability is linearly decodable from hidden states, but it is unclear
which of its forms share a representation and whether the signal is useful in
dialogue. We contribute a turn-labeled multi-turn benchmark (423 conversations,
1{,}661 labeled turn-states) and an evaluation harness with a simulated user
who answers clarifying questions, and use them with six datasets and six
open-weight LLMs to test how far probes for unanswerability carry. Probes
transfer robustly between datasets that share a ground of unanswerability:
missing information in math (AUROC 0.77--0.97) and in a passage
(SQuAD~2.0$\leftrightarrow$MuSiQue, 0.77--0.90). Probes for epistemic
``known-unknowns'' transfer poorly to math, but this separation weakens under
lexical controls and changes with layer and coordinate system, so it remains
unresolved. Single-turn probes fail zero-shot to detect when a conversation
becomes answerable; in-structure probes recover it, but no better than a
bag-of-words classifier. A gate on the calibrated probe, with no model
fine-tuning, fires on underspecified turns far more precisely than chance, and
its end-task success comes within 0.08 of a gate given the true labels. Yet
across four models it does not reliably beat vanilla generation or prompted
consolidation. The remaining gap lies mostly in how models use a
clarification, not in detection.
\end{abstract}

%% file: sections/introduction.tex
\section{Introduction}

Recent work shows that whether a question is answerable is linearly decodable
from a language model's hidden states, and uses this to mitigate abstention
failures in reasoning models~\citep{liu2025abstention}. That work is, however,
demonstrated within a single domain (unanswerable math), and its outlook is
training-centric: ``Future work aims to explore training-time alignment
strategies to improve abstention fidelity''~\citep{liu2025abstention}.
Transfer beyond the training data has been examined before: answerability
probes carry across extractive QA benchmarks~\citep{slobodkin2023curious},
while hidden-state error and truthfulness detectors generalize only partially
across tasks and domains~\citep{orgad2025llms,liu2024universal}. What remains
open is \emph{which} forms of unanswerability share a representation. A
question can be unanswerable because the question omits information (a math
problem missing a quantity), because the supplied context omits it (a passage
without the answer), or because no one knows the answer. All three call for
the same behavior, abstention, but need not share a latent direction usable
as a read-out without fine-tuning the model.

Independently, LLMs degrade sharply when instructions are revealed
incrementally across an under-specified conversation rather than stated all at
once~\citep{laban2025lost}. That work maps the degradation and its
workarounds but leaves its scope open (``Establishing whether models get
lost in conversation in other languages, or in tasks that involve multiple
modalities [\ldots] could help establish the scope of the degradation observed
in LLM multi-turn capabilities''~\citep{laban2025lost}), and it does not provide
a mechanism that \emph{detects}, turn by turn, when a conversation is still
under-specified.

These two open problems share a premise: if the latent unanswerability signal
were domain- and structure-general, it could gate generation without any
fine-tuning of the model: read the probe, then abstain on unanswerable inputs and ask a
clarifying question in under-specified dialogue. We evaluate RQ1 on six
open-weight models from four families and test the full premise end-to-end on
the four 8--12B models. It holds only in parts:
\begin{itemize}
    \item \textbf{RQ1, across datasets.} The signal transfers robustly between
    datasets that share a ground of unanswerability. Whether epistemic
    unknowability is represented separately remains unresolved: its apparent
    separation depends on layer, lexical controls and coordinate system.
    \item \textbf{RQ2, across structure.} Single-turn probes do not transfer
    zero-shot to turn-level dialogue states at the layers where they are
    strongest. Turn-level answerability is recoverable in-structure, but no
    better than a TF-IDF classifier on the user turns recovers it.
    \item \textbf{RQ3, at deployment.} A leakage-free gate fires on
    underspecified turns far more precisely than chance, and its end-task
    success comes close to that of a gate given the true labels. It still does
    not beat prompting workarounds.
\end{itemize}

\paragraph{Contributions.}\footnote{Code, frozen datasets, the decision-rule record, and all result
tables will be released publicly.}
\begin{itemize}
    \item \textbf{A turn-labeled multi-turn underspecification benchmark.} The
    benchmark has 423 conversations and 1{,}661 turn-states, with constraints
    revealed one per turn and labels fixed by construction. It was audited by
    eight annotators and repaired by a judge from a model family other than
    the generator's, under rules fixed before the repair ran.
    \item \textbf{An answering-user evaluation harness} for clarification
    policies. The simulated user answers from the not-yet-revealed constraints
    but cannot leak anything beyond them. Oracle, position and rate-matched
    random gates separate detector error from intervention cost.
    \item \textbf{Careful transfer and deployment results, mostly negative.}
    \begin{itemize}
        \item \textbf{RQ1:} probes transfer robustly within a ground of
        unanswerability: math 0.77--0.97, and extractive QA 0.77--0.90,
        replicating \citealp{slobodkin2023curious}. The known-unknowns
        separation is fragile under lexical, layer and coordinate controls
        (Tables~\ref{tab:transfer} and~\ref{tab:kunk}).
        \item \textbf{RQ2:} single-turn probes fail zero-shot on turn-states,
        and in-structure probes only match TF-IDF (Table~\ref{tab:multiturn}).
        \item \textbf{RQ3:} the probe gate beats vanilla on single-turn
        abstention but trails a one-line prompt. In dialogue, its firing
        precision is 0.74--0.88, against 0.54--0.58 for a random gate, and its
        success is within 0.08 of the oracle gate. Its end-task endpoint,
        fixed in advance, is still negative (Table~\ref{tab:intervention}).
    \end{itemize}
\end{itemize}

%% file: sections/method.tex
\section{Method}\label{sec:method}

\subsection{Probing for Unanswerability}
For every (model, dataset) pair we run a single prompt-only forward pass per
example (the model's own chat template, no generation), cache all layers'
hidden states in fp16, and read out at the last \emph{prompt} token (a
mean-pooled read-out is cached alongside; last-token is used throughout).
Probes are $\ell_2$-regularized logistic regressions on the read-out at one
layer: features are standardized by a scaler fit on the \emph{training data
only}; the regularization strength is chosen from a log-grid
($10^{-4}$--$10^{2}$) by inner cross-validation on the training fold;
within-domain numbers are 5-fold stratified CV over seeds $\{13, 42, 77\}$
and reported as mean~$\pm$~sd. Cross-domain and cross-structure evaluations
apply the \emph{frozen} probe with its frozen train-domain scaler to the
entire test set: no test-side fitting of any kind. Layer indices are
model-specific. For the transfer matrix, the common layer maximizes mean
within-domain cross-validation AUROC. For LODO, we evaluate the full layer
sweep and report the best held-out-domain layer as an explicitly optimistic,
descriptive upper bound; it is not used to fit or tune the probe.

We use six open-weight instruction models from four families (Qwen3.5-9B,
Qwen3.6-27B, gemma-4-12B-it, gemma-4-31B-it, Ministral-3-8B-Instruct, and
granite-4.1-8b), plus DeepSeek-R1-Distill-Qwen-7B as the anchor for
reproducing the single-domain result of \citet{liu2025abstention} (reproduced at AUROC
0.871/0.948 on SUM/UMWP, within the published range). The two 27--31B models
enter RQ1 only: RQ2 and RQ3 require multi-turn roll-outs and many generation
runs per model, which our compute budget allowed for the four 8--12B base
models. Label-shuffle nulls and TF-IDF baselines on identical splits are run
for all six models and reported inline with the results. Nulls are at chance
throughout. On the math and extractive sets the probe beats TF-IDF by a wide
margin; the known-unknowns sets are lexically separable, which we analyze
separately.

\subsection{Domains and Datasets}
Six frozen single-turn datasets (600 balanced items each; SUM 568), two per ground of
unanswerability:
\begin{itemize}
    \item \emph{information missing from the question} (math):
    SUM~\citep{song2025sum}, minimally edited answerable/unanswerable pairs of
    math problems, and UMWP~\citep{sun2024umwp}, unanswerable math word
    problems;
    \item \emph{information missing from the context} (extractive):
    SQuAD~2.0~\citep{rajpurkar2018squad2}, and MuSiQue~\citep{trivedi2022musique}
    built as minimal pairs in the setting of \citet{slobodkin2023curious}.
    Each MuSiQue question appears in both classes; its unanswerable context
    replaces the missing hop's supporting paragraph with the most similar
    paragraph that contains no hop answer;
    \item \emph{epistemic unknowability} (known-unknowns):
    SelfAware~\citep{yin2023selfaware} (e.g. ``Did we invent math or
    discover it?''), and KUQ~\citep{amayuelas2024kuq}, restricted to its
    epistemic categories (future, unsolved, controversial, counterfactual).
\end{itemize}
Labels follow each source. Every frozen set is versioned with a content hash
and deduplicated against the others.

\subsection{Transfer and Geometry}
RQ1 measures the full train$\times$test matrix: for each ordered domain pair,
the frozen probe (with its train-domain scaler) is evaluated on the full test
domain; the diagonal is within-domain CV. A leave-one-domain-out (LODO)
variant trains one pooled probe on all-but-one domain and tests on the
held-out domain. Geometry compares class-separation directions
$\Delta\mu = \overline{x}_{y=1} - \overline{x}_{y=0}$ per (domain, seed,
fold) at the common best layer: pairwise cosines, the participation-ratio
effective rank of the stacked directions, and the principal angle between
the \{math, extractive\} subspace and the known-unknowns axis, all
mean~$\pm$~sd over the 15 seed$\times$fold splits. The robustness analyses
add three things:
\begin{itemize}
    \item bootstrap-over-items 95\% CIs with within-dataset split-half null
    angles, plus a whitened (shrinkage-LDA) variant in a common pooled
    coordinate system;
    \item a lexical residualization, which removes the component of the hidden
    state predicted from TF-IDF features, with the map fit on training folds
    only;
    \item layer selection from the training dataset alone.
\end{itemize}
Each analysis has a decision rule committed before it ran
(Appendix~\ref{app:prereg}).

\subsection{A Turn-Labeled Multi-Turn Benchmark}
Following the sharded-instruction setting of \citet{laban2025lost}, each
conversation reveals the atomic constraints of a task one per user turn in a
randomized order; a deterministic rule labels every \emph{turn-state} (the
conversation up to and including user turn $t$, before the model's reply)
as underspecified (1) until the predefined sufficiency set of
constraints has been revealed, then answerable (0). Sources: grade-school
math (numeric answer), small Python functions (reference implementation),
and data-to-text (slot-value sentence); 21\% of conversations are
early-fully-specified distractors so that position alone does not determine
the label. The generated pool (504 conversations) was audited by eight human
annotators (728 verdicts, disjoint shards plus a shared calibration set),
which surfaced a systematic defect: default-confirming constraints padded
into sufficiency sets. A shrink-only repair by a judge model from a family
other than the generator's, with rules fixed in advance and validated against
the annotators' calibration set, plus a rule-based drop pass, froze the
benchmark at
\textbf{423 conversations / 1{,}661 turn-states} (drops: 58
question-arrives-last math items, 19 unexplained flags, 4 judge conflicts).
For RQ2, each model first replies to every turn of every conversation
(greedy roll-out), so turn-state read-outs contain the model's own
intermediate replies in context; an excluded-replies variant is the control.

\subsection{Probe-Gated Intervention}
RQ3 turns the probe into a gate that requires no fine-tuning of the model
(a worked example of one conversation, from constraints to gate decisions,
is in Appendix~\ref{app:worked}). The raw probe score at the
last prompt token is mapped to a calibrated probability by Platt
scaling~\citep{platt1999} fit on held-out \emph{training-side} validation
data, and thresholded at $\tau = 0.5$ ($\tau$ swept in the appendix). Gate
probes never see gated items: single-turn items of domain $D$ are gated by
the LODO probe trained on the other domains (transfer is the mechanism);
multi-turn conversations are gated out-of-fold: probes trained on
grouped conversation folds that never contained the gated conversation,
since single-turn probes do not transfer across structure
(see the RQ2 results). When the gate fires, the model does not answer:
single-turn, it emits the same fixed abstention string as the prompted
abstain baseline; multi-turn, a fixed prompt appended to the flagged turn
makes the same model ask one clarifying question about the most important
missing constraint. Baselines on identical, N-matched items: vanilla
generation, prompted ask-when-unsure, prompted abstain, and RECAP-style
consolidation (a recap turn restating all user turns before the final
answer)~\citep{laban2025lost}. All decoding is greedy with a 384-token cap.
In multi-turn runs the user \emph{answers} clarifying questions. An LLM
(\texttt{gpt-5.6-terra}, T=0, from a family distinct from the generator and
the evaluated models) only \emph{selects} which not-yet-revealed constraint
answers the model's question. The user then sends that constraint verbatim;
the math goal sentence is also selectable. A generic question receives the
next scripted constraint, and an unanswerable question receives a fixed ``no
further details'' reply. Scripted turns already revealed are skipped, and the
same user serves every condition.

To separate detector error from intervention cost, three reference gates use
the same clarify prompt: an \emph{oracle} gate (fires iff the sufficiency set
is unrevealed), a \emph{position} gate (fires while any scripted constraint is
unrevealed), and a \emph{random} gate that matches the probe gate's
per-conversation firing rate (seeds 13/42/77).
End-of-conversation success is objective per source: numeric match (math),
differential execution against the reference implementation on a typed
input battery (code), and slot-value coverage (data-to-text); single-turn
correct behavior is refusal on unanswerable items and a normalized gold
match on answerable ones. Paired comparisons use exact McNemar tests and
paired bootstrap CIs on matched items.

%% file: sections/results.tex
\section{Results}

Before each analysis we committed its decision rule to version control, then
ran it. Appendix~\ref{app:prereg} lists the commits and timestamps, and the
code release will include each report as it stood at its commit. This
is a timestamped record kept by the authors, not registration with a third
party, and one rule was amended before its run; we flag both where relevant.

\subsection{RQ1: Cross-Dataset Transfer}\label{sec:rq1}

Within-dataset probes are strong everywhere (mean diagonal 0.94--0.98;
Table~\ref{tab:transfer}). The robust transfer result concerns datasets that
share a \emph{ground} of unanswerability. The two math sets transfer at
0.77--0.97 in all six models, in both directions. The two extractive sets
also transfer in both directions in every model: SQuAD~2.0 and MuSiQue,
the latter built as minimal pairs in the setting of
\citet{slobodkin2023curious}, transfer at 0.77--0.90 at the six-dataset
common layer. This replicates \citeauthor{slobodkin2023curious}'s cross-dataset
result on current models.

\begin{figure*}[t]
\centering
\includegraphics[width=\textwidth]{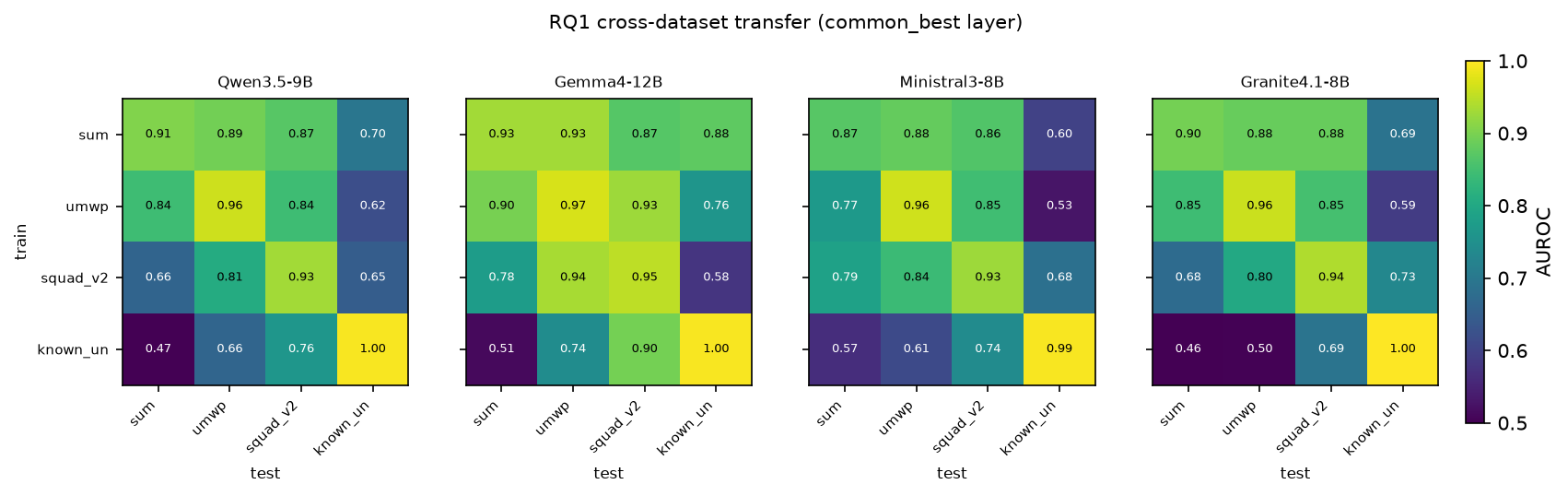}
\caption{Train$\times$test transfer matrices on the four core datasets for
the four 8--12B models (AUROC, mean over seeds). All six models on all six
datasets are in Appendix Figure~\ref{fig:heatmap6x6}.}
\label{fig:heatmap}
\end{figure*}

Transfer \emph{across} grounds depends on model and layer. At the layer that
maximizes within-domain AUROC over the four core datasets, math and
extractive QA transfer at 0.66--0.94 (Figure~\ref{fig:heatmap}). Adding the
two new datasets moves the common layer in three of six models (Qwen3.5
20$\rightarrow$29, Granite 26$\rightarrow$31, Gemma4-12B 32$\rightarrow$31).
At that layer the same math$\leftrightarrow$SQuAD cells span 0.38--0.94, and
math$\leftrightarrow$MuSiQue spans 0.17--0.90 (Appendix
Figure~\ref{fig:heatmap6x6}). Known-unknowns
probes are at or near chance on SUM (SelfAware$\rightarrow$SUM 0.46--0.57)
despite $\approx$0.995 within-dataset AUROC.

By the decision rule we fixed before the first transfer run (cross-dataset mean $\le 0.65$ or drop $\ge 0.20$
$\Rightarrow$ dissociation), four of six models land on dissociation. Two of
these verdicts sit on the boundary under seed-and-sampling bootstrap
(Qwen3.5 0.93, Ministral 0.87 of replicates; Table~\ref{tab:transfer}). If the
layer is instead chosen per \emph{training} dataset, so that no target label
touches selection, all six models are classified as dissociation. These
verdicts concern the four-dataset matrix as a whole; the next paragraph asks
which part of it is robust.

\begin{table}[t]
\centering
\small
\setlength{\tabcolsep}{2.2pt}
\resizebox{\columnwidth}{!}{%
\begin{tabular}{lccccc}
\toprule
 & \multicolumn{3}{c}{four datasets (common layer)} & \multicolumn{2}{c}{six datasets} \\
model & diag & cross [95\% CI] & drop [95\% CI] & SQ$\rightarrow$MQ & MQ$\rightarrow$SQ \\
\midrule
Qwen3.5-9B    & 0.948 & 0.732 [.721,.743] & 0.216 [.195,.236] & 0.88 & 0.81 \\
Qwen3.6-27B   & 0.979 & 0.834 [.825,.843] & 0.145 [.129,.159] & 0.88 & 0.77 \\
Gemma4-12B    & 0.962 & 0.808 [.799,.818] & 0.153 [.138,.169] & 0.89 & 0.83 \\
Gemma4-31B    & 0.976 & 0.757 [.747,.767] & 0.219 [.202,.235] & 0.90 & 0.90 \\
Ministral3-8B & 0.939 & 0.727 [.715,.738] & 0.213 [.191,.233] & 0.84 & 0.86 \\
Granite4.1-8B & 0.949 & 0.717 [.706,.729] & 0.232 [.214,.249] & 0.88 & 0.83 \\
\bottomrule
\end{tabular}}
\caption{RQ1 transfer (AUROC, mean over seeds $\{13,42,77\}$). Left: the
four-core-dataset matrix at the common best layer. Diag is within-dataset;
cross averages all off-diagonal cells (including SUM$\leftrightarrow$UMWP);
drop = diag $-$ cross. CIs come from a seed and sampling bootstrap. Right:
extractive-pair transfer between SQuAD~2.0 (SQ) and MuSiQue (MQ) at the
six-dataset common layer. Per-model LODO and source-only-layer results are in
the appendix.}
\label{tab:transfer}
\end{table}

\paragraph{Is known-unknowns represented separately?}
SelfAware is lexically separable (TF-IDF 0.94), and so is the second
known-unknowns benchmark we add, KUQ (0.94). Both build the answerable class
from factoid QA sets and write the unanswerable class separately, so the label
coincides with the source genre. A TF-IDF classifier trained on either set
transfers to the other at 0.89--0.91. KUQ therefore \emph{replicates}
SelfAware rather than controlling for its confound. We tested the separation
three ways, each with a decision rule committed before the run
(Table~\ref{tab:kunk}).

\begin{itemize}
    \item \textbf{Replication on KUQ.} At the six-dataset layer, KUQ's probe
    also transfers poorly to SUM ($\le 0.65$ in 5/6 models). At the four-dataset
    layer, KUQ$\rightarrow$SUM exceeds 0.65 in 3/6.
    \item \textbf{Lexical residualization.} We remove the component of the
    hidden state that TF-IDF features predict, fitting the projection on
    training folds only. A known-unknowns signal survives: within-dataset
    AUROC is 0.81--0.92 for SelfAware, and items that TF-IDF misclassifies are
    separated at 0.98--0.99. Its \emph{non-transfer} to math does not survive:
    SelfAware$\rightarrow$SUM rises above 0.65 in 3/6 models, which our
    decision rule classifies as negative. Our committed rule compared
    residualized SelfAware$\leftrightarrow$KUQ transfer with \emph{raw}
    known-unknowns$\rightarrow$math transfer and held in 6/6. Compared
    like for like, both residualized, it holds in only 4/6: it fails for both
    Gemma models, and Qwen3.5 passes by 0.02.
    \item \textbf{Geometry with uncertainty.} In $\Delta\mu$ space the
    \{math, extractive\} subspace is further from the known-unknowns direction
    than math is from extractive in 6/6 models (CI $> 0$), well above the
    within-dataset split-half floor (14--29$^{\circ}$). After whitening by the
    pooled covariance (shrinkage LDA), the gap is positive in no model and
    significantly negative in 3/6. With MuSiQue as the extractive set, the raw
    gap reverses for both Gemma models.
\end{itemize}

\begin{table}[t]
\centering
\small
\setlength{\tabcolsep}{2pt}
\resizebox{\columnwidth}{!}{%
\begin{tabular}{lccccc}
\toprule
 & SA$\rightarrow$SUM & KUQ$\rightarrow$SUM & SA$\leftrightarrow$KUQ / & \multicolumn{2}{c}{$D$ (deg)} \\
model & raw / res. & L$_6$ / L$_4$ & SA$\rightarrow$math, res. & $\Delta\mu$ & whitened \\
\midrule
Qwen3.5-9B    & .47 / .61 & .52 / .68 & .66 / .64 & $+5.2^*$ & $-2.1^*$ \\
Qwen3.6-27B   & .55 / \textbf{.77} & .65 / .65 & .81 / .79 & $+8.2^*$ & $-0.4$ \\
Gemma4-12B    & .51 / \textbf{.80} & \textbf{.74} / \textbf{.75} & \underline{.72 / .86} & $+5.8^*$ & $-1.8^*$ \\
Gemma4-31B    & .52 / \textbf{.79} & .45 / .45 & \underline{.78 / .86} & $+6.2^*$ & $-2.4^*$ \\
Ministral3-8B & .57 / .63 & .60 / .60 & .72 / .70 & $+15.9^*$ & $+0.6$ \\
Granite4.1-8B & .46 / .53 & .59 / \textbf{.71} & .78 / .53 & $+10.0^*$ & $+1.4$ \\
\bottomrule
\end{tabular}}
\caption{Robustness of the known-unknowns separation (SA = SelfAware).
Transfer is AUROC; res.\ means after removing the TF-IDF-predictable
component of the hidden state. KUQ$\rightarrow$SUM is shown at the six-dataset
(L$_6$) and four-dataset (L$_4$) common layers. The fourth column compares
residualized SA$\leftrightarrow$KUQ transfer (mean of both directions) with
residualized SA$\rightarrow$\{SUM, UMWP\}; underlined where known-unknowns
transfers no better within its ground than to math. $D$ is the angle between
the \{math, extractive\} subspace and the known-unknowns direction minus the
math--extractive angle; $^*$: 95\% bootstrap CI excludes 0. Bold: above the
0.65 threshold.}
\label{tab:kunk}
\end{table}

The known-unknowns separation is therefore unresolved. Raw probes and raw
$\Delta\mu$ geometry suggest a separate direction in every model, but removing
lexical cues, whitening the coordinates, or changing the layer or the
extractive dataset each removes it in some models. We read RQ1 as robust
transfer \emph{within} a ground of unanswerability, and make no claim about
separation \emph{between} grounds.

\emph{Controls}: label-shuffle nulls are 0.47--0.52 for all six models and six
datasets. The model-independent TF-IDF baseline trails the probe by 44--72
AUROC points on the math and extractive sets; SUM's and MuSiQue's minimal
pairs defeat it outright (0.24, 0.48). Best-layer selection cannot create the
weak cross-ground transfer: the weakest headline AUROC survives Bonferroni
correction across layers (appendix).

\subsection{RQ2: Cross-Structure Transfer to Multi-Turn}\label{sec:rq2}

Can the single-turn probe detect, zero-shot, the moment a conversation
becomes answerable? No -- and the failure is itself structured
(Table~\ref{tab:multiturn}). Applied to turn-states, frozen single-turn
probes are at or \emph{below} chance at the mid/late layers where they live
(0.20--0.55 for three of four models): the single-turn underspecification
direction anti-correlates with turn-level underspecification. What
zero-shot signal exists sits in the first one or two layers (0.77--0.85) and
is partially surface-driven. Behaviorally, the frozen probe declares
``sufficient'' one to two turns \emph{prematurely} in 74--98\% of
conversations (mean lag $-1.3$ to $-2.3$ turns): it would stop asking
clarifying questions too early.

\begin{table}[t]
\centering
\small
\resizebox{\columnwidth}{!}{%
\begin{tabular}{lcccc}
\toprule
model & zero-shot & ceiling & within-turn & $\angle$ to ST \\
\midrule
qwen3.5-9b     & 0.784 & 0.927 & 0.76--0.84 & 73.9$^{\circ}$ \\
ministral-3-8b & 0.774 & 0.935 & 0.79--0.86 & 56.7$^{\circ}$ \\
granite-4.1-8b & 0.789 & 0.925 & 0.76--0.84 & 81.2$^{\circ}$ \\
gemma-4-12b    & 0.845 & 0.934 & 0.79--0.86 & 58.7$^{\circ}$ \\
\bottomrule
\end{tabular}}
\caption{RQ2 turn-level AUROC. Zero-shot: best layer of the frozen
single-turn probe (mid/late layers are at or below chance, 0.20--0.55).
Ceiling: probe trained on turn-states with conversation-grouped CV.
Within-turn: ceiling restricted to fixed turn indices 2--4, min--max over
turns (absolute position is fixed; relative position still varies with
conversation length). $\angle$ to ST: principal angle of the multi-turn $\Delta\mu$ to
the whole single-turn subspace.}
\label{tab:multiturn}
\end{table}

Turn-level underspecification is nonetheless strongly decodable
\emph{in-structure}: probes trained on turn-states (grouped by conversation
so no conversation straddles train/test) reach 0.925--0.935 AUROC. Because
turn-level labels are position-correlated by construction (relative
position alone scores 0.934), we also evaluate \emph{within} fixed turn
indices, where absolute position is constant: the ceiling stays at
0.76--0.86 (min--max over turns 2--4 and models; Table~\ref{tab:multiturn}),
while zero-shot retains only a weak trace
(0.51--0.79). Within a turn index, however, the label remains partly
predictable from the text of the user turns alone: a TF-IDF classifier on
them, trained with the same conversation-grouped splits, reaches 0.79
within-turn (0.91 overall), on par with the ceiling averaged over turns 2--4
(0.79--0.81 across models). We
therefore read the in-structure result as showing that the information is
present and linearly available, not that the probe captures more than the
surface text does. Labels also remain largely predictable from the
conversation's design within a turn index. Position relative to the total
number of turns, which a turn-state prefix does not reveal, matches the probe
within 0.02. A logistic model on design features (turn index, total turns,
source task, context length) exceeds it by 0.07--0.10. The probe's within-turn signal is therefore consistent with, but
not proof of, the model inferring task size from content. This makes the zero-shot failure starker: a signal that
a bag-of-words classifier recovers is not picked up by the single-turn
direction. Label-shuffle nulls for the ceiling are clean
(0.486--0.496). Geometrically, the multi-turn direction is rotated
57--81$^{\circ}$ from the \emph{entire} single-turn subspace (raw
$\Delta\mu$ coordinates), and its cosine to the known-unknowns direction is
negative in all four models. Two further controls:
per-domain zero-shot transfer follows the RQ1 ordering (math probes
transfer best, known-unknowns worst); and removing the model's own replies
from the context \emph{improves} zero-shot transfer for granite (0.829 vs
0.789) while slightly shrinking the rotation: the model's own reply
tokens are part of what rotates the direction. Replies truncated at the
384-token cap (17--63\% of replies, depending on the model) shift
within-turn AUROC by up to 0.10 for the ceiling and 0.15 for zero-shot, in
opposite directions across models (Appendix~\ref{app:trunc}). These shifts
are far smaller than the zero-shot failure itself.

\subsection{RQ3: Probe-Gated Abstention and Clarification}\label{sec:rq3}

RQ3 asks whether the signal, deployed as a leakage-free calibrated gate,
improves end-task performance. We test it on all four 8--12B models
(Table~\ref{tab:intervention}).

\begin{table}[t]
\centering
\small
\setlength{\tabcolsep}{3pt}
\resizebox{\columnwidth}{!}{%
\begin{tabular}{lcccc}
\toprule
condition & Qwen3.5 & Granite & Gemma4-12B & Ministral \\
\midrule
\multicolumn{5}{l}{\emph{multi-turn success, answering user}} \\
vanilla               & 0.764 & 0.598 & 0.506 & 0.811 \\
ask-when-unsure       & 0.759 & 0.811 & 0.459 & 0.766 \\
consolidation (RECAP) & \textbf{0.837} & \textbf{0.825} & 0.631 & 0.856 \\
\textbf{probe gate} ($\tau{=}0.5$) & 0.693 & 0.681 & 0.584 & 0.853 \\
\quad oracle gate (true labels) & 0.721 & 0.681 & 0.660 & 0.861 \\
\quad random gate, rate-matched (3 seeds) & 0.510 & 0.537 & 0.300 & 0.584 \\
\quad position gate$^\dagger$ & 0.820 & 0.719 & \textbf{0.754} & \textbf{0.915} \\
premature: vanilla $\rightarrow$ probe & .98$\rightarrow$.43 & .99$\rightarrow$.62 & .94$\rightarrow$.23 & .92$\rightarrow$.69 \\
probe firing precision / recall & .88 / .75 & .86 / .70 & .74 / .90 & .85 / .68 \\
random firing precision & .57 & .57 & .55 & .58 \\
\midrule
\multicolumn{5}{l}{\emph{single-turn correct behavior}} \\
vanilla        & 0.378 & 0.378 & 0.427 & 0.315 \\
abstain prompt & \textbf{0.636} & \textbf{0.611} & \textbf{0.663} & \textbf{0.582} \\
probe gate     & 0.589 & 0.525 & 0.609 & 0.515 \\
\bottomrule
\end{tabular}}
\caption{RQ3 on N-matched items (423 conversations; 2{,}368 single-turn items).
Multi-turn success is objective per source: numeric match, differential code
execution, or slot coverage. The user simulator answers clarifying questions
from the not-yet-revealed constraints. $^\dagger$The position gate fires while
any scripted constraint is unrevealed; it needs the script length and is not
deployable. Firing precision is the share of fired turns that are
underspecified, and recall the share of underspecified turns on which the gate
fires; random precision is the mean over three seeds (oracle: 1/1 by
construction). The scripted-user results and the cost-sensitive gate are in the
appendix.}
\label{tab:intervention}
\end{table}

\paragraph{Single-turn.} In all four models the gate beats vanilla by
+0.15 to +0.21 (McNemar $p \le 10^{-24}$), but it trails the one-line abstain
\emph{prompt} by 0.05--0.09. The cause is threshold transfer, which RQ1
predicts. For Qwen and Granite, which our offline risk--coverage analysis
covers, the leave-one-domain-out gate's calibrated threshold fires on 94--99\%
of math items and on none of SQuAD. Yet its ranking \emph{within} each dataset
beats chance in 7 of 8 model--dataset pairs, and at the same abstention budget
an oracle detector would beat the prompt. A full
$\tau$-sweep (appendix) rescues neither model, and a cost-sensitive
threshold fixed in advance does not help either.

\paragraph{Multi-turn, with a user who answers.} A scripted user who cannot
answer clarifying questions makes asking pure cost. We therefore evaluate with a
simulated user
in the style of CLAM~\citep{kuhn2022clam}, who knows the not-yet-revealed
constraints. An LLM only \emph{selects} which hidden constraint answers the
model's question, and the user then replies with that constraint verbatim, so
nothing beyond the scripted information can leak. The simulator was validated
before use: on 60 real exchanges it was judged correct in 56 by a human, with
no over-reveals.

The endpoint, fixed before any gated run was scored, is \textbf{negative}. The
probe gate beats vanilla significantly in two of four models (Granite
$+0.083$, Gemma $+0.078$; Holm-adjusted $p < 0.02$). It is significantly
\emph{worse} for Qwen ($-0.071$) and indistinguishable for Ministral. It loses
to RECAP-style consolidation in two of four ($-0.144$ for Qwen and Granite).
The reference gates locate the gap:
\begin{itemize}
    \item \textbf{Detection.} Against the turn labels, the probe gate fires
    with precision 0.74--0.88, against 0.54--0.58 for a random gate. It
    misses 10--32\% of underspecified turns (recall 0.68--0.90; the oracle
    is exact by construction), and these misses account for 68--100\% of its
    premature answers. These premature answers cost little: conversations
    with one end correct as often as those without, within 0.04 in every
    model, because the model revises its answer once the remaining
    constraints arrive (Appendix~\ref{app:worked}). This is why better
    detection would add little: the true-label gate's success is at most 0.08
    higher.
    \item \textbf{Asking at the right moment matters.} At the same firing
    rate, the probe gate beats randomly timed asking by +0.14 to +0.28 in
    every model, and random asking is strongly harmful.
    \item \textbf{The remaining limit is how the model uses a clarification.}
    For Qwen, even the true-label gate does not beat vanilla. The
    non-deployable position gate keeps asking until the script is exhausted,
    and it is the best gate in every model. This suggests that more persistent
    asking can pay off.
\end{itemize}
Letting the user answer raises the probe gate by +0.10 (Qwen) and +0.19
(Granite) over the scripted-user setting, so a user who cannot answer gives a
loose lower bound. In short, the probe gate asks at better-than-chance moments
and nearly matches a true-label gate, but neither turns asking into end-task
gains over consolidation. Firing precision and recall are a descriptive
analysis added after the endpoint was scored.

%% file: sections/related_work.tex
\section{Related Work}

\paragraph{Probing internal states for answerability and truthfulness.}
Linear structure in the residual stream has been reported for truthfulness:
supervised probes separate true from false statements~\citep{azaria2023internal},
unsupervised consistency objectives recover latent truth
directions~\citep{burns2023dlk}, and true/false datasets exhibit emergent
linear geometry that supports steering~\citep{marks2023geometry}. Closest to
us, \citet{liu2025abstention} show that \emph{answerability} is linearly
decodable from hidden states of reasoning models and use it to analyze
abstention failures -- within a single (math) domain.
\citet{slobodkin2023curious} probe answerability on SQuAD~2.0, Natural
Questions and MuSiQue and find that the learned subspace transfers across
these extractive benchmarks; \citet{orgad2025llms} show that hidden-state
error detectors do not transfer universally across ten tasks, and
\citet{liu2024universal} search for a single truthfulness hyperplane shared
across tasks and domains. We build on this line but vary the \emph{ground}
of unanswerability rather than the benchmark: information missing from the
question, information missing from the context, and epistemic
unknowability. We then follow the same signal across conversational
structure (RQ2) and into a deployed gate (RQ3). Transfer is robust within a
ground, while the separation between grounds remains unresolved.

\paragraph{Abstention and selective prediction.}
Abstention in LLMs is typically elicited by prompting, fine-tuning, or
confidence heuristics; \citet{wen2024abstentionsurvey} survey the space and
explicitly pose whether abstention can be ``a meta-capability that
transcends specific tasks or domains.'' Our unanswerable sets instantiate
distinct abstention grounds: information missing from the question
(SUM~\citep{song2025sum}, UMWP~\citep{sun2024umwp}), information missing
from a passage (SQuAD~2.0~\citep{rajpurkar2018squad2}), and questions
unanswerable for anyone (SelfAware~\citep{yin2023selfaware},
KUQ~\citep{amayuelas2024kuq}). Our RQ1 results give a partial answer to the
survey's question. Missing-information abstention transfers across
datasets that share a ground. Whether epistemic unknowability shares that
representation is not settled by our data: its apparent separation weakens
under lexical controls and changes with layer and coordinate system.

\paragraph{Multi-turn underspecification and clarifying questions.}
\citet{laban2025lost} document that models ``get lost'' in sharded,
underspecified conversations and evaluate prompt-level workarounds (recap
and snowball-style consolidation), which we adopt as baselines. We add what
that line lacks: a benchmark whose \emph{turn-states are labeled} for
answerability by construction (with human audit and rule-based repair),
a latent detector evaluated turn-by-turn against those labels, and a gate
that asks a clarifying question exactly when the detector fires. Our RQ3
result sharpens the workaround comparison: consolidation attacks context
loss and wins. Detection-gated asking fires far more precisely than chance
and nearly matches a true-label gate, yet does not convert this into end-task
gains, even with a user who answers.
Selective clarification has also been studied with simulated users who hold
the missing information: CLAM~\citep{kuhn2022clam} asks only for ambiguous
inputs and answers from such a simulator, and \citet{zhang2025clarify}
decide \emph{when} a clarification is worth its cost from its expected
benefit. Our gate makes this decision from a hidden-state probe rather than
from the model's own outputs.

\paragraph{Calibration and confidence.}
Our gate reads a single scalar through Platt scaling~\citep{platt1999} fit
on training-side validation data only, a monotone recalibration that
leaves probe ranking untouched; the model itself is never fine-tuned. Throughout, we control the classic confound of
confidence read-outs: label-shuffle nulls and TF-IDF baselines on identical
splits separate representation content from lexical surface, and a
position-only baseline plus within-turn evaluation do the same for
conversational position (RQ2 results).

%% file: sections/conclusion.tex
\section{Conclusion}

We asked how far the latent unanswerability signal in LLM hidden states
carries, using a new turn-labeled benchmark and an answering-user harness.
\begin{itemize}
    \item \textbf{Across datasets.} Probes transfer robustly when two datasets
    share a ground of unanswerability. Whether epistemic unknowability is
    represented separately remains open: the evidence changes with layer,
    lexical controls and coordinate system.
    \item \textbf{Across structure.} Single-turn probes fail zero-shot on
    turn-level dialogue states. In-structure probes recover turn-level
    answerability, but no better than a classifier on the user turns.
    \item \textbf{At deployment.} A leakage-free gate fires on underspecified
    turns far more precisely than chance, and comes within 0.08 of a
    true-label gate in end-task success. It still does not beat consolidation.
\end{itemize}
These are mostly negative results, reported with their controls. The
benchmark, the simulator and the result tables are the parts we expect to be
most useful, and we release all of them.

%% file: sections/limitations.tex
\section*{Limitations}

\textbf{Scale and architecture coverage.} Four model families and two size
tiers (8--12B and 27--31B), but only Qwen and Gemma are represented in both;
the Qwen pair also changes model generation (3.5 to 3.6), so it is not a
controlled scaling comparison. granite-4.1-8b is a hybrid Mamba-2/transformer, so layer
indices are not comparable across architectures; all layer selections are
per-model. \textbf{Synthetic multi-turn data.} The benchmark's conversations
are generated, then human-audited and repaired under a fixed
protocol (728 verdicts surfaced a systematic sufficiency-set defect; the
repaired, frozen set drops 81 of 504 conversations); labels are deterministic
given sufficiency sets, but the conversations are not naturally occurring
dialogue. \textbf{Correlational probes.} All probe results are decoding
claims, not causal ones: we did not run the planned activation-steering
ablation, so nothing here shows the probed direction causally mediates
abstention. \textbf{Layer selection and multiple comparisons.} Best layers come from
full layer sweeps; the appendix quantifies the exposure: the weakest
headline AUROC survives Bonferroni correction across all layers
($p \approx 2\times10^{-13}$ vs $\alpha \approx 10^{-3}$), the RQ2 ceiling
is layer-insensitive (0.885--0.934 across all layers), and held-out LODO
layer selection biases transfer upward, so it cannot create the weak transfer
we report.
\textbf{Annotators are authors.} The eight benchmark annotators are authors of
this paper, so the audit is not independent. Its role was to find defects,
which it did; the labels themselves follow from the sufficiency sets by rule.
\textbf{Simulated user.} The answering user replies only with scripted
constraints, selected by an LLM, and never improvises or misunderstands. It
was validated on 60 exchanges by a single annotator. Real users would answer
less cleanly, which affects every ask-based condition alike.
\textbf{Known-unknowns datasets.} Both known-unknowns benchmarks build their
answerable class from a different source than their unanswerable class, so
label and genre coincide. Lexical residualization reduces this confound but
cannot remove it; a lexically matched known-unknowns set would be needed.
\textbf{Run recovery.} Some RQ3 generation runs were resumed from
checkpoints after infrastructure failures, and one run spans two GPU types;
all are listed in a run log released with the code.
\textbf{Judge-scored clarify quality} was declared in advance
as secondary-only evidence and is not reported; all multi-turn success
metrics are objective (numeric match, differential code execution, slot
coverage). \textbf{Token accounting} uses whitespace tokens, identical
across conditions, so only deltas are meaningful.

\section*{Ethical Considerations}

\textbf{Potential risks.} A probe gate that fires too often withholds answers
users need. One that fires too rarely gives false assurance that a model
``knows'' when it cannot answer. Our gates do not beat simple prompting
baselines, so we do not recommend deploying them as a safeguard. The probes
read hidden states of open-weight models and offer no defense against
adversarial inputs.

\textbf{Data and annotation.} All data are public research datasets, used under
their licenses, or synthetic conversations without personal information. The
eight benchmark annotators are authors of this paper and took part voluntarily;
no data about them were collected.

\textbf{Use of AI assistants.} Beyond the LLMs that are part of our method
(benchmark generation, benchmark repair, and the user simulator;
Section~\ref{sec:method}), we used AI assistants to help write code, to draft
parts of the text, and to fix grammar. The authors designed the study, fixed
the decision rules before each run, checked the results, and take
responsibility for all content.

%% file: sections/appendix.tex
\appendix

\section{Decision Rules Committed Before Each Run}\label{app:prereg}

Table~\ref{tab:prereg} lists each decision rule that the paper describes as
fixed in advance. For each it gives the commit that fixed the rule and the
earliest timestamp of the run's output: the first result row, or the first
API spend-log entry for judge runs. The code release will
include a folder \texttt{prereg/} with each report exactly as it stood at that commit,
with status \emph{planned}, the rule written, and no results. Every result
row also records the git hash of the code that produced it.

This record establishes order, not independence. It is kept in our own
repository rather than a third-party registry. For several rules the run
followed within minutes, because the code was ready and only the rule
remained to be written. Reports not listed here were written together with
their results, and we do not describe their rules as fixed in advance.

\begin{table}[h]
\centering
\small
\setlength{\tabcolsep}{3pt}
\resizebox{\columnwidth}{!}{%
\begin{tabular}{lcll}
\toprule
rule & commit & committed (UTC) & first result (UTC) \\
\midrule
RQ1 dissociation rule & \texttt{3600fc0} & 07-11 21:32 & 07-11 21:44 \\
LODO pooled probe & \texttt{c577bc9} & 07-12 11:34 & 07-12 11:35 \\
Direction geometry & \texttt{462b8dd} & 07-12 16:12 & 07-12 16:33 \\
Benchmark repair protocol & \texttt{bc2049f} & 07-17 14:23 & 07-17 14:25 \\
RQ3 endpoint + metrics & \texttt{d80bc2c} & 07-20 09:49 & 07-20 10:20 \\
\midrule
Robustness datasets, text baselines & \texttt{1bd8f6d} & 09-25 07:07 & 09-25 07:07 \\
Simulator validation + RQ3-v2 endpoint & \texttt{cf5450f} & 09-25 08:09 & 10-01 07:31 \\
RQ2 text/metadata baselines & \texttt{a824f63} & 09-25 09:33 & 09-25 09:34 \\
RQ3 offline + cost-sensitive gate & \texttt{281609b} & 09-25 09:34 & 09-25 09:39 \\
Table~\ref{tab:transfer} CIs; source-only layers & \texttt{ac7e547} & 09-25 09:34 & 09-25 09:35 \\
Lexical residualization; geometry CIs & \texttt{6740551} & 10-01 08:08 & 10-01 08:17 \\
Second known-unknowns / extractive sets & \texttt{ece0f8f} & 10-01 08:38 & 10-01 08:57 \\
RQ3-v2 scoring (Holm, best baseline) & \texttt{96705a2} & 10-01 23:10 & 10-02 02:35 \\
\bottomrule
\end{tabular}}
\caption{Rule commits and first outputs (all 2026, UTC). Rows below the line
cover the robustness analyses and the answering-user evaluation. The two-sided headline (unified mechanism vs
dissociation) is fixed in the project specification committed on 07-10. The
answering-user scoring commit restates the \texttt{cf5450f} endpoint and makes two
implicit details explicit: Holm correction across models, and the per-model
best prompted baseline. Before that commit, exactly one generation output
(Granite vanilla) had been inspected, as a harness sanity check. The rule for
the second known-unknowns set was amended by the authors before its run, to
read SelfAware$\leftrightarrow$KUQ transfer only after lexical
residualization.}
\label{tab:prereg}
\end{table}

\section{Threshold Sensitivity}

Table~\ref{tab:tau} sweeps the gate threshold $\tau$ over the single-turn
suite offline: for each $\tau$, items whose recorded calibrated gate
probability exceeds $\tau$ receive the abstention string, all others the
vanilla reply (valid because decoding is greedy on identical prompts; the
gated runs record a gate probability for every item). qwen improves
monotonically toward high $\tau$ (0.603 at $\tau{=}0.9$) but never reaches
the abstain prompt (0.636); granite peaks near $\tau{=}0.2$--$0.3$ (0.549,
prompt 0.611). For multi-turn, the recorded final-turn probabilities give
the would-fire rate at the already-sufficient final turn: for qwen 19.1\%
at $\tau{=}0.5$ $\rightarrow$ 6.9\% at 0.7 $\rightarrow$ 1.9\% at 0.8; for
granite 7.8\% $\rightarrow$ 0\% at 0.7 (earlier-turn trajectories change
with $\tau$, so full multi-turn curves require re-generation and are left
to future work).

\begin{table}[h]
\centering
\small
\begin{tabular}{lccccc}
\toprule
model & $\tau$ & correct & abst$_{\text{unans}}$ & false-abst & acc$_{\text{ans}}$ \\
\midrule
qwen    & 0.1 & 0.575 & 0.888 & 0.726 & 0.262 \\
qwen    & 0.3 & 0.582 & 0.869 & 0.676 & 0.294 \\
qwen    & 0.5 & 0.589 & 0.861 & 0.643 & 0.317 \\
qwen    & 0.7 & 0.600 & 0.849 & 0.585 & 0.351 \\
qwen    & 0.9 & 0.603 & 0.781 & 0.475 & 0.424 \\
granite & 0.1 & 0.548 & 0.841 & 0.730 & 0.254 \\
granite & 0.3 & 0.549 & 0.803 & 0.663 & 0.294 \\
granite & 0.5 & 0.526 & 0.726 & 0.586 & 0.325 \\
granite & 0.7 & 0.492 & 0.601 & 0.454 & 0.383 \\
granite & 0.9 & 0.441 & 0.383 & 0.237 & 0.499 \\
\bottomrule
\end{tabular}
\caption{Single-turn $\tau$ sweep (pooled, $n{=}2368$): correct behavior,
abstention on unanswerable, false abstention on answerable, accuracy on
answerable. Full grid (nine values of $\tau$) in
\texttt{results/rq3\_tau\_single.csv}.}
\label{tab:tau}
\end{table}

\section{Multi-Turn Dataset Datasheet}\label{app:datasheet}

\textbf{Language and coverage.} All datasets and conversations are in
English. The benchmark covers three task types (arithmetic word problems,
small Python functions, and data-to-text records), and its data-to-text
records are invented.

\textbf{Construction.} 504 generated conversations over three sources
(grade-school math, small Python functions, data-to-text records); atomic
constraints revealed one per user turn in randomized order; deterministic
turn-state labels from predefined sufficiency sets; 21\% distractors
fully specified before the final turn. \textbf{Audit.} Eight annotators,
728 verdicts (disjoint 59-item shards + a shared 32-item calibration set):
flags concentrated on math/code (26.6\%/23.0\% vs 0.4\% data-to-text) and
converged on one defect -- default-confirming constraints padded into
sufficiency sets. \textbf{Repair and freeze.} A shrink-only repair by a
non-generator-family judge model (fixed prompt, temperature 0), validated
against the calibration set under a rule fixed before the repair ran; conversations
whose flags the repair could not explain, whose decomposition conflicted,
or whose answerability flip is carried by the final question utterance were
dropped (58 math + 19 + 4). Frozen set: \textbf{423 conversations, 1{,}661
turn-states} (code 164 / data-to-text 167 / math 92), content-hashed.
\textbf{Roll-outs.} Each model replies to every turn (greedy, 384-token
cap); truncation rates are a model-verbosity signature (17--63\%) and are
recorded per manifest.

\textbf{Annotation.} The eight annotators are authors of this paper and
took part voluntarily. No personal data were collected.

\textbf{Compute.} Activation extraction and generation used NVIDIA V100
32\,GB GPUs (fp16) and, for the 27--31B models and some generation runs,
H100 GPUs (bf16). Including development and failed runs, we estimate about
200 GPU-hours in total, roughly 85\% of them on V100s. The largest share,
about 100 GPU-hours, went to the answering-user evaluation. Probing,
transfer, geometry and all statistics run on CPU from cached activations. API
costs were about \$62 for the user simulator and under \$10 for generating
and repairing the benchmark.

\textbf{Software.} Probes, scalers, Platt calibration and TF-IDF baselines use
scikit-learn 1.9 with the settings in Section~\ref{sec:method}. Models run
through Hugging Face transformers with greedy decoding. Code success is
scored in a CPython 3.14 sandbox. Exact versions are pinned in the
\texttt{pyproject.toml} of the code release.

\section{Layer Selection and Multiple Comparisons}

Best layers are chosen from full per-layer sweeps ($L$ = 33--65 for RQ1;
33--49 for the four-model RQ2 evaluation). This raises two questions.
\textbf{(1) Is any headline
number a sweep artifact?} A conservative Hanley--McNeil bound on the
\emph{weakest} headline AUROC (within-turn ceiling, 0.686 at turn~1,
$n{=}423$) gives $z \approx 7.2$, $p \approx 2\times10^{-13}$ -- far below
the Bonferroni-corrected threshold $\alpha = 0.05/49 \approx 10^{-3}$;
every stronger cell passes a fortiori. The RQ2 in-structure ceiling is
additionally layer-\emph{insensitive}: 0.885--0.934 across \emph{all}
non-embedding layers of all four models, so no layer choice drives it.
\textbf{(2) Does optimistic layer selection bias the transfer conclusion?}
The LODO curves of the four 8--12B models peak narrowly (1--4 layers within 0.02 of
best), but the
selection direction is conservative for our claim: choosing the best layer
per held-out domain can only inflate transfer, so it cannot create the weak
cells we report. The
common-best-layer transfer matrix (Table~\ref{tab:transfer}, left) uses
within-dataset CV selection only. Every reported LODO value, including the
27--31B extension, is explicitly treated as an optimistic upper bound.

\section{Additional Ablations and Controls}

\textbf{Excluded-replies read-out.} Removing the model's own replies from
turn-state contexts improves zero-shot transfer for granite (0.829 vs
0.789) and slightly shrinks the single-turn-to-multi-turn rotation
(78.6$^{\circ}$ vs 81.2$^{\circ}$): the model's own reply tokens are part
of what rotates the direction. \textbf{Position controls.} Relative
position alone ($-t/T$) reaches 0.934 AUROC on turn-state labels, so RQ2
claims are also stated within fixed turn indices. Within a turn index
absolute position is constant, but relative position is not (conversations
have 3--5 turns): $-t/T$ still scores 0.70/0.88/0.86 at turns 2/3/4, and
a metadata model that knows the conversation length reaches 0.89. That
length is not observable from a turn-state prefix; the observable metadata
model (position, constraints revealed, context length, source) reaches
0.68, and a TF-IDF classifier on the user turns 0.79
(\texttt{results/rq2\_baselines.csv}). \textbf{Shuffle nulls.} Ceiling label-shuffle nulls
0.486--0.496; single-turn nulls 0.47--0.52 across all six models and four
datasets, run with TF-IDF baselines on identical splits at each
dataset's own best layer (\texttt{results/controls.csv}).
\textbf{Per-seed tables} are in the released CSVs
(\texttt{results/}), each row carrying config hash and seed; the
27--31B extension follows the same seeds and probe configuration, and its
transfer, pooled, geometry, and control rows are in the same CSVs.
\textbf{Infrastructure.} Hidden-state extraction and generation ran in
fp16 on NVIDIA V100 (32~GB) nodes for the 8--12B models and in bf16 on a
single 80~GB H100 for the 27--31B pair; extraction dtype is recorded in
each cache's metadata. Probing, statistics, and evaluation are CPU-only
(PyTorch, transformers, and scikit-learn; exact versions pinned in the
released lockfile).

\section{Worked Example}\label{app:worked}

Table~\ref{tab:worked} traces one benchmark conversation
(\texttt{mt-math-0004}) through the answering-user evaluation for qwen3.5-9b
with the probe gate. The task (``Maria bakes 6 trays of 12 muffins, donates a
quarter, sells the rest at \$2; how much does she raise?'', gold answer \$108)
is decomposed into four atomic constraints, all in the sufficiency set. At
each turn the last-prompt-token read-out at the gate layer is scored by the
out-of-fold probe and mapped through Platt scaling. The gate fires at
$p \ge 0.5$ and appends the fixed clarify prompt; the simulated user then
answers from the constraints not yet revealed.

The trace shows both behaviors behind the RQ3 results. At turns 1--2 the gate
fires and the user answers each question with the constraint it asks for, in
the order the model asks, not the script's. At turn 3 the gate misses an
underspecified turn ($p = 0.35$), and the model answers prematurely with a
revenue of \$144. The donation constraint then arrives as the next scripted
turn, and the model revises its answer to the correct \$108. Across models,
68--100\% of the probe gate's premature answers occur on turns where it did
not fire, as here. The rest are turns where it fired but the model answered
anyway.

\begin{table*}[t]
\centering
\small
\begin{tabular}{cp{4.6cm}cccp{6.4cm}}
\toprule
$t$ & user turn (source) & label & $p$ & gate & model reply (excerpt) \\
\midrule
1 & ``\ldots I made 6 trays of them.'' (script) & 1 & 0.995 & ask & ``\ldots I need to know how many muffins are on each tray. How many muffins are on each tray?'' \\
2 & ``Each tray holds 12 muffins.'' (answer) & 1 & 0.793 & ask & ``\ldots the most important piece of information missing is the selling price per muffin \ldots How much are you planning to sell each muffin for?'' \\
3 & ``The muffins I do sell go for \$2 each.'' (answer) & 1 & 0.345 & answer & ``\ldots Total muffins: $6 \times 12 = 72$ \ldots Total potential revenue: 72 muffins $\times$ \$2 \ldots'' (premature) \\
4 & ``Before selling any, I set aside one quarter \ldots to donate \ldots'' (script) & 0 & 0.403 & answer & ``\ldots $\frac{1}{4}$ of 72 = 18 muffins \ldots the amount you will actually raise \ldots is \$108.'' \\
\bottomrule
\end{tabular}
\caption{Worked example (qwen3.5-9b, probe gate, answering user). Source:
\emph{script} is the next scripted constraint; \emph{answer} is the simulated
user's reply to the model's question. Label: 1 = underspecified (sufficiency
set not yet revealed). $p$: Platt-calibrated gate probability.}
\label{tab:worked}
\end{table*}

\section{Transfer Matrices for All Six Models}

\begin{figure*}[t]
\centering
\includegraphics[width=0.92\textwidth]{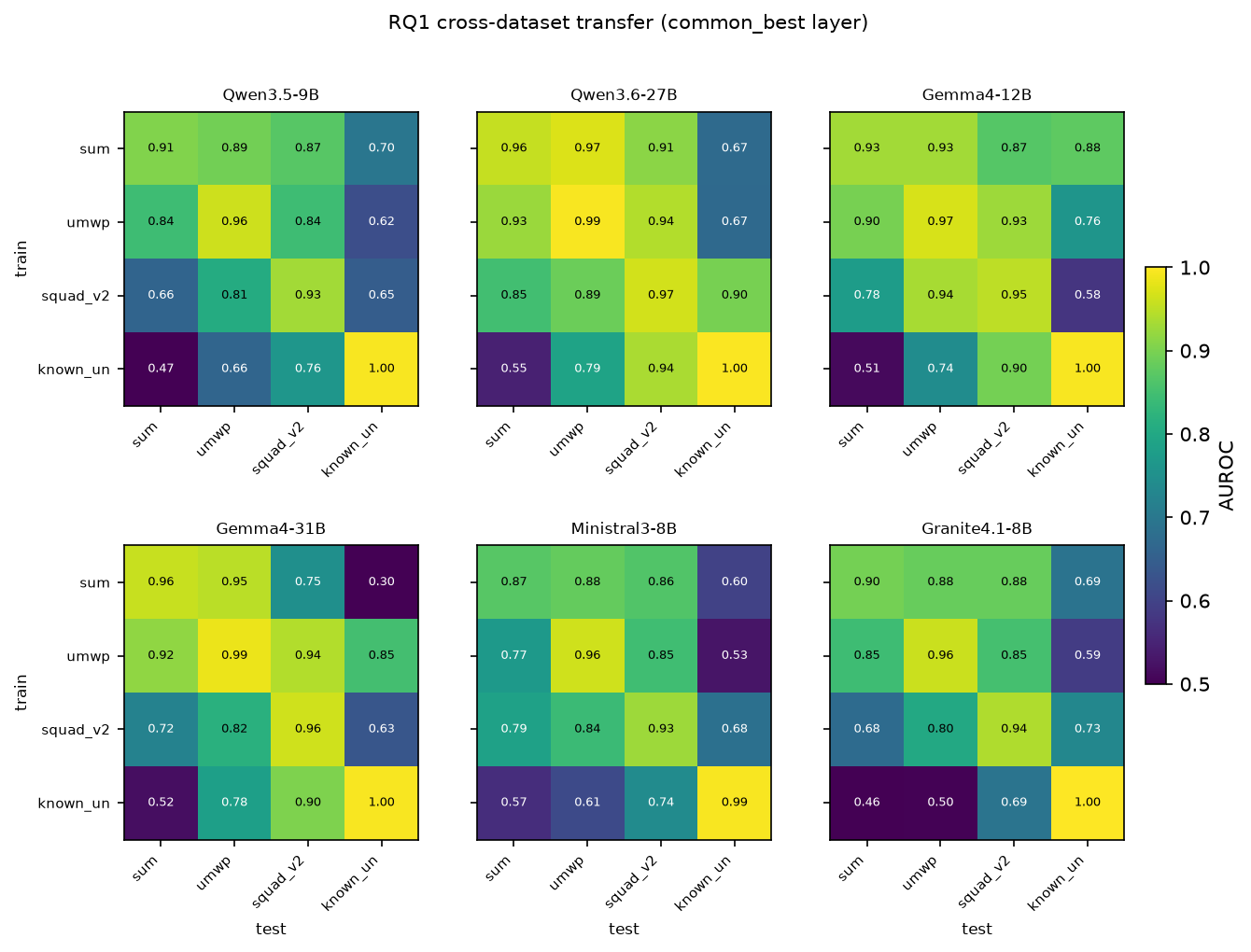}
\caption{Train$\times$test transfer matrices at each model's common best layer
(AUROC, mean over seeds $\{13,42,77\}$) for all six models; the color scale
is clipped at 0.5. Rows are the training dataset, columns the test dataset.
Generated from \texttt{results/transfer\_bestlayer.csv} by
\texttt{scripts/plot\_transfer\_heatmap.py}.}
\label{fig:heatmap6}
\end{figure*}

\section{Pooled and Source-Only Transfer}\label{app:lodo}

Table~\ref{tab:lodo} reports, on the four core datasets, a probe pooled
over all but one domain and evaluated frozen on the held-out domain. It also
shows the dissociation verdict when each training dataset's layer is chosen
from that dataset alone, so that no target label touches layer selection.

\begin{table}[h]
\centering
\small
\setlength{\tabcolsep}{3pt}
\resizebox{\columnwidth}{!}{%
\begin{tabular}{lccc|cc}
\toprule
 & \multicolumn{3}{c|}{LODO held-out} & \multicolumn{2}{c}{source-only layers} \\
model & math & extr & k-unk & cross / drop & verdict \\
\midrule
Qwen3.5-9B    & 0.649 & 0.878 & 0.888 & 0.680 / 0.272 & dissoc. \\
Qwen3.6-27B   & 0.796 & 0.936 & 0.850 & 0.766 / 0.213 & dissoc.\ (was mixed) \\
Gemma4-12B    & 0.731 & 0.920 & 0.856 & 0.720 / 0.244 & dissoc.\ (was mixed) \\
Gemma4-31B    & 0.780 & 0.939 & 0.812 & 0.670 / 0.307 & dissoc. \\
Ministral3-8B & 0.671 & 0.869 & 0.680 & 0.682 / 0.261 & dissoc. \\
Granite4.1-8B & 0.665 & 0.900 & 0.770 & 0.721 / 0.231 & dissoc. \\
\bottomrule
\end{tabular}}
\caption{Left: LODO, a descriptive upper bound reported at the best
held-out-domain layer. Right: transfer when each training dataset's layer is
chosen from that dataset alone. Selecting the layer on seed 13 and evaluating
on seeds 42 and 77 gives the same 6/6.}
\label{tab:lodo}
\end{table}

\section{RQ3: Scripted User and Cost-Sensitive Gate}\label{app:rq3more}

Table~\ref{tab:scripted} contrasts a scripted user, who cannot answer clarifying questions, with the answering user
of Section~\ref{sec:rq3}. It also reports the cost-sensitive
gate. Its threshold is derived in closed form from a fixed cost matrix
(an extra exchange costs $\kappa{=}0.1$; asking on an already-sufficient turn adds $\lambda{=}0.7$) and the
model's answerable accuracy: $\tau^* = (\kappa+\lambda-1+\mathrm{acc})/(\mathrm{acc}+\lambda)$.

\begin{table}[h]
\centering
\small
\setlength{\tabcolsep}{3pt}
\resizebox{\columnwidth}{!}{%
\begin{tabular}{lcccc}
\toprule
 & Qwen3.5 & Granite & Gemma4-12B & Ministral \\
\midrule
vanilla, scripted user      & 0.759 & 0.589 & -- & -- \\
probe gate, scripted user   & 0.596 & 0.492 & -- & -- \\
vanilla, answering user     & 0.764 & 0.598 & 0.506 & 0.811 \\
probe gate, answering user  & 0.693 & 0.681 & 0.584 & 0.853 \\
cost gate, answering user   & 0.669 & 0.690 & 0.407 & 0.856 \\
\midrule
user turns, vanilla$\rightarrow$probe & 4.3$\rightarrow$4.8 & 4.0$\rightarrow$4.9 & 4.1$\rightarrow$4.9 & 4.2$\rightarrow$5.3 \\
``no further details'' $\ge$1, probe & 0.71 & 0.62 & 0.68 & 0.77 \\
\bottomrule
\end{tabular}}
\caption{Multi-turn success under the scripted user (two models) and the answering user. The cost-sensitive gate does not improve on
$\tau{=}0.5$. The last row is the share of probe-gated conversations in which
the model asked at least one question that the hidden constraints could not
answer.}
\label{tab:scripted}
\end{table}

\section{RQ2: Reply Truncation}\label{app:trunc}

Model replies in the RQ2 roll-outs are capped at 384 tokens, and the cap is
hit by 16.6\% (Qwen), 30.7\% (Granite), 39.3\% (Ministral) and 63.0\% (Gemma)
of replies. Table~\ref{tab:trunc} splits within-turn AUROC (turns 2--4) by
whether any earlier reply in the context was truncated. Under our
threshold fixed before the split was run ($|\Delta| \ge 0.05$) truncation is material, with a
model-specific sign. It does not account for the zero-shot failure, which
puts mid/late layers at or below chance.

\begin{table}[h]
\centering
\small
\begin{tabular}{lcc}
\toprule
model & $\Delta$ ceiling & $\Delta$ zero-shot \\
\midrule
Qwen3.5-9B     & $+0.069 \pm 0.010$ & $+0.071$ \\
Ministral3-8B  & $+0.101 \pm 0.007$ & $+0.066$ \\
Granite4.1-8B  & $+0.025 \pm 0.009$ & $-0.050$ \\
Gemma4-12B     & $-0.005 \pm 0.025$ & $-0.153$ \\
\bottomrule
\end{tabular}
\caption{$\Delta$ = AUROC(clean context) $-$ AUROC(context with a truncated
reply), within turns 2--4; mean $\pm$ sd over seeds. The zero-shot probe is
seed-deterministic.}
\label{tab:trunc}
\end{table}

\section{Transfer Matrices on Six Datasets}

\begin{figure*}[t]
\centering
\includegraphics[width=0.92\textwidth]{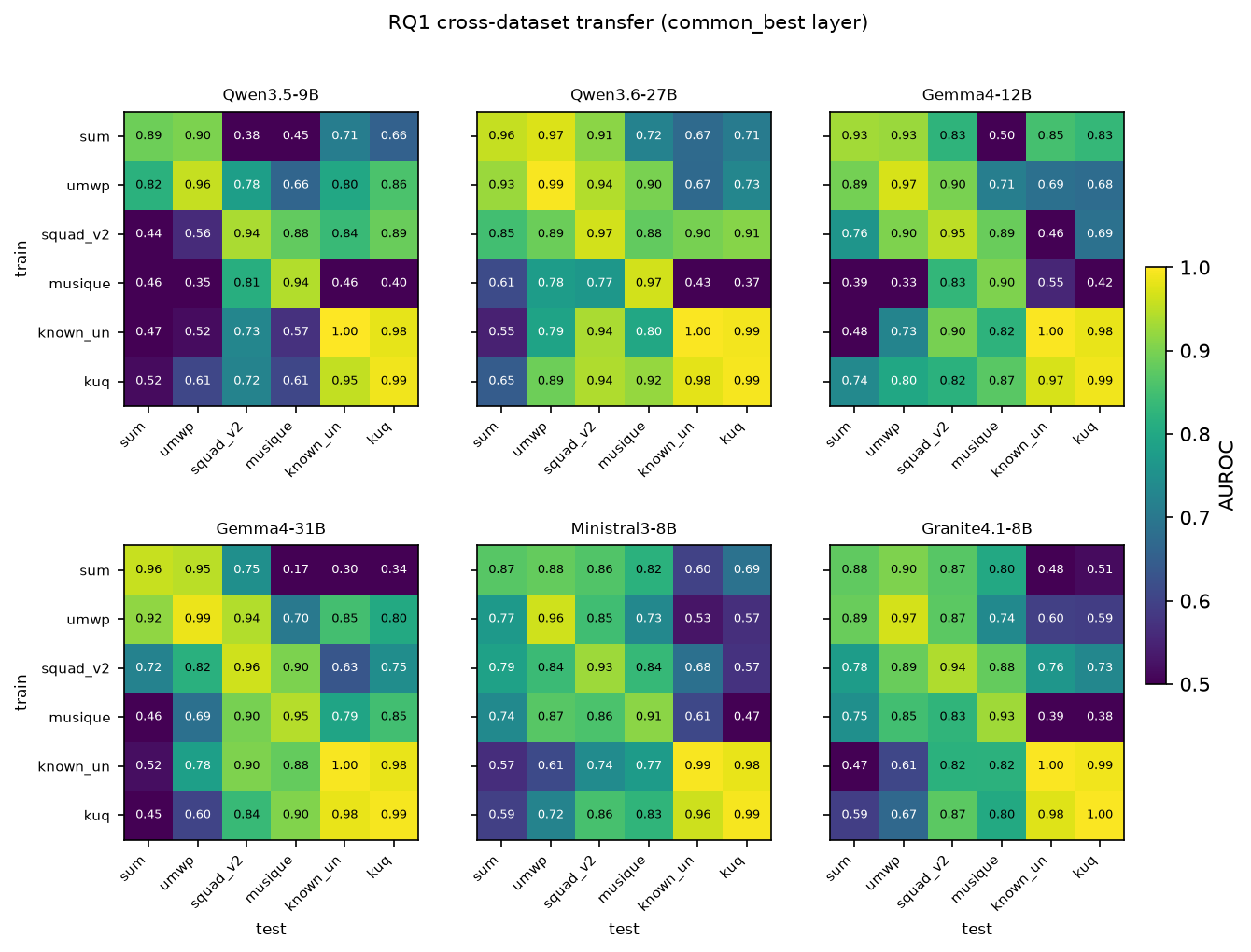}
\caption{Train$\times$test transfer on all six datasets at each model's
six-dataset common best layer (AUROC, mean over seeds $\{13,42,77\}$; color
clipped at 0.5). Generated from \texttt{results/rq1\_v2/transfer\_bestlayer.csv}
by \texttt{scripts/plot\_transfer\_heatmap.py}.}
\label{fig:heatmap6x6}
\end{figure*}

Figure~\ref{fig:heatmap6x6} shows the full train$\times$test transfer
matrices for all six models on all six datasets, including MuSiQue and KUQ.

\section{Annotator Instructions}\label{app:instructions}

The full text given to the eight annotators, all authors of this paper, is
reproduced below. It accompanied a self-contained HTML annotation tool that
showed each conversation with its task, the computed turn labels and, on
request, the gold answer. Essential turns had a blue border and optional turns
an orange one. Each annotator received a disjoint shard plus a shared
calibration set (Appendix~\ref{app:datasheet}).

\begin{quote}
\small
\textbf{Annotator instructions --- multi-turn underspecification}

Thanks for helping verify our benchmark. You have one file,
\texttt{annotate\_<your-letter>.html} (alpha, beta, \ldots). Open it in any
browser --- no install, works offline, and your progress saves automatically.

\textbf{What you're checking.} Each item is a short conversation where the
user reveals the task \textbf{one piece per turn}. We've auto-labeled every
turn as:
\begin{itemize}
    \item \textbf{UNDERSPECIFIED} --- the task can't be answered yet
    (information still missing), or
    \item \textbf{ANSWERABLE} --- enough has been said to answer.
\end{itemize}
Your job: \textbf{is that label correct at every turn?}

The rule: a turn is answerable \textbf{once all the ESSENTIAL constraints
have appeared} (blue-bordered turns). \textbf{OPTIONAL} turns (orange) are
redundant extra detail --- they must \textbf{not} change whether the task is
answerable. So the label should switch from UNDERSPECIFIED to ANSWERABLE
exactly at the turn that completes the essential information, and stay
ANSWERABLE afterward.

\textbf{How to judge each conversation.}
\begin{enumerate}
    \item Read the \textbf{task} and turns. Reveal the \textbf{gold answer} if
    it helps.
    \item Check the flip point (marked ``$\blacktriangle$ becomes answerable at
    turn N''): is it right?
    \begin{itemize}
        \item Too early --- it says answerable but something essential is
        still missing.
        \item Too late --- it's answerable but still marked underspecified.
        \item An ``optional'' turn actually adds needed information (so it
        wasn't optional).
        \item The task itself is broken, nonsensical, or a later turn leaks
        earlier.
    \end{itemize}
    \item Press \textbf{\texttt{a}} if all labels are correct, or
    \textbf{\texttt{f}} to \textbf{flag} it and type a one-line note on what's
    wrong (which turn, and why).
\end{enumerate}
Keys: \texttt{a} correct $\cdot$ \texttt{f} flag $\cdot$ \texttt{j}/\texttt{k}
previous/next.

\textbf{When you're done.} Click \textbf{Export JSON}, save
\texttt{annotations\_<your-letter>.json}, and send it back. You'll see
$\sim$91 conversations; some are shared across annotators on purpose (that's
how we measure agreement). When in doubt, flag with a note rather than
guessing --- a flagged item gets a second look, an unflagged one is trusted.
\end{quote}